\documentclass[runningheads]{llncs}
\usepackage{graphicx}
\usepackage{subfigure, caption}
\usepackage{epstopdf}
\begin{document}
\title{Identification of Abnormality in Maize Plants From UAV Images Using Deep Learning Approaches}
\titlerunning{Abnormality Identification in Maize Plants Using Deep Learning}
%
\author{Aminul Huq\inst{1}\orcidID{0000-0002-6500-6097} \and
Dimitris Zermas\inst{2}\orcidID{0009-0006-0412-2218} \and
George Bebis\inst{1}\orcidID{0009-0000-6222-5967}
}
\authorrunning{A. Huq et al.}
%
\institute{University of Nevada Reno, Reno NV 89512, USA \and
Sentera
\email{aminul.huq@nevada.unr.edu,dimitris.zermas@sentera.com,bebis@unr.edu}\\
}
\maketitle              
\begin{abstract}

Early identification of abnormalities in plants is an important task for ensuring proper growth and achieving high yields from crops. Precision agriculture can significantly benefit from modern computer vision tools to make farming strategies addressing these issues efficient and effective. As farming lands are typically quite large, farmers have to manually check vast areas to determine the status of the plants and apply proper treatments. In this work, we consider the problem of automatically identifying abnormal regions in maize plants from images captured by a UAV. Using deep learning techniques, we have developed a methodology which can detect different levels of abnormality (i.e., low, medium, high or no abnormality) in maize plants independently of their growth stage. The primary goal is to identify anomalies at the earliest possible stage in order to maximize the effectiveness of potential treatments. At the same time, the proposed system can provide valuable information to human annotators for ground truth data collection by helping them to focus their attention on a much smaller set of images only. We have experimented with two different but complimentary approaches, the first considering abnormality detection as a classification problem and the second considering it as a regression problem. Both approaches can be generalized to different types of abnormalities and do not make any assumption about the abnormality occurring at an early plant growth stage which might be easier to detect due to the plants being smaller and easier to separate. As a case study, we have considered a publicly available data set which exhibits mostly Nitrogen deficiency in maize plants of various growth stages. We are reporting promising preliminary results with an 88.89\% detection accuracy of low abnormality and 100\% detection accuracy of no abnormality. 

\keywords{Precision Agriculture  \and Deep Learning \and Computer Vision \and Maize Plants  \and Abnormality Detection.}
\end{abstract}
\section{Introduction}
The negative impact of climate change and it's repercussions related to the agricultural sector and the environment is increasing at an alarming rate. Additionally, with the increased demand of food because of the ever-increasing population, the agricultural sector faces huge problems in the future. With the help of artificial intelligence and different methodological approaches, these problems may be handled to some extend. Automation of different agricultural tasks reduces human effort, increases food production and mitigates the adverse effects on the environment to some extend. Since most of the agricultural farms encompass a vast amount of land and produce, using sophisticated technological approaches has a significant impact in providing ease to human efforts and the environment. Machine learning and its applications have aided in various research and practical problems. Many times, machine learning approaches have surpassed human performance. Examples include face recognition, question answering, medical image analysis, speech recognition and others \cite{b1,b2,b3,b4}. It's application in the field of precision agriculture is also noteworthy \cite{b5,b6}. 

In order to ensure food security and economic stability for the U.S., production of corn plays a vital role. The U.S. is one of the major corn producers in the world. According to the data collected from the United States Department of Agriculture (USDA) in 2022, corn farmers of the U.S. produced about 13.7 billion bushels of corn at 79.2 million acres of land \cite{b7}. Apart from using this commodity as a food source, it is also used as a bio-fuel and other industrial applications. Proper inspection for any abnormalities in the plant leaves and subsequent treatment have a huge impact on the yield of maize plants. In the initial phase of an abnormality, the leaves of the plants usually turn from healthy green to different shades of yellow. There are various reasons for this like a disease or nutrient deficiency. There are also cases where the leaves turn yellow due to changes of weather, lack of water or other reasons. In general, manual assessment of the plants and their leaves is required from an expert who traverses the whole area where the plants are cultivated. This is a tedious process which needs to be done multiple times during the lifetime of the plants. 

There have been several attempts to automate these processes, however, a general approach irrespective of the growth stage of the plants or type of abnormality is still under investigation. Past challenges include low resolution images, inefficient data collection approach, and problem-depended methodology based on specific deficiencies, viruses, insects \cite{b8,b9,b21}. Lack of a large set of properly labelled data is also another major limitation which prevents researchers from training effective deep learning models. 

Using powerful sensors, Unmanned Aerial Vehicles (UAVs) are capable of taking high resolution images from a large area of a maize field leading to an efficient data collection approach for abnormality inspection purposes but also for collecting large amounts of data for training deep learning models\cite{b10}. The goal of this research work is to develop a system which can identify and quantify the level of abnormality from images irrespective of the abnormality type or level (i.e., low, medium, high or none). There are two main benefits behind the proposed system. First, it can provide useful information to farmers in terms of potential plant abnormalities and their quantification. Second, it can provide valuable assistance to human annotators for ground truth collection by focusing their attention on a subset of images instead of examining all the images captured in a UAV data collection flight. The main contributions of this research work can be summarized as follows:
\begin{itemize}
    \item A methodology which can be used to identify and quantify abnormalities in maize plants. The proposed methodology can be extended to different types of plants.  
    \item A customized EfficientNet-B0 is used as a baseline model in this study. Comparison with other models are shown, demonstrating the superiority of the baseline model.
    \item The labelled images with annotations which have been created for this research work are released of the benefit for the research community\footnote[1]{https://github.com/aminul-huq/Abnormality-Corn-ISVC-23}.
\end{itemize}

The rest of the paper is organized as follows: Section 2 provides a review of related works. Details about the data set and how it was prepared is mentioned in Section 3. The proposed methodology, and the baseline deep learning architecture can be found in Section 4. Section 5 presents the experiments performed and discusses our results. Finally, Section 6 provides our conclusions and directions for future research.
\section{Background}
Identifying nutrient deficiencies and abnormalities is a major concern in precision agriculture research. There are some interesting research works that have been performed in the past. Chore and Thankachan \cite{b11} attempted to identify Potassium, Nitrogen, Copper, Zinc and Phosphorus deficiencies from the leaves of orange, cotton, apple, banana, mango, litchi, henna, gooseberry, and okra plants. The plant leaves were manually collected from the fields using a multi-frequency visible light leaf scanning approach. Using multi-frequency analysis coupled with image processing, 16 features were used to train a deep learning model. The proposed approach involved a four step lengthy process, however, the results reported were quite high. The collected data was not made publicly available. 

Rahadiyan et al. \cite{b12} performed similar work on Chilli plants. In particular, they extracted texture and color based feature and used a Multi-Layer Perceptron model to identify seven classes namely healthy, Phosphor, Magnesium, Sulfur, Calcium, Magnesium-Sulfur, and Multi-deficiency. However, the size of the total dataset was only 817. Moreover, the dataset included only images of individual plant leaves which were picked and placed on a white background rather than the whole plant in the field. Using the combination of RGB, Greyscale and LBP features the authors were able to achieve only 79.67\% of accuracy.   

To figure out the stress caused to plants by drought, Tejasri et al. \cite{b13} used a UAV to capture aerial RGB images of maize plants and utilized an ensemble model based on U-Net and U-Net++ where ResNet34 was used as the backbone of the model. The UAV device captured images of the whole maize field from where the authors extracted 150 samples of healthy and stressed crops. The authors created a segmentation mask using Otsu's method and a naive thresholding approach. They were able to improve overall performance by stacking and averaging the performance of U-Net and U-Net++ models. The final mIoU score of the ensemble model was 0.7163. As the segmentation mask or the ground truth was based on the Otsu's and naive thresholding method it can be assumed that the ground truth was not fine tuned enough to capture all the details in the images. A polygon based method may improve the ground truth collection process and the segmentation models performance. Using image processing techniques for detecting Nitrogen deficiency in rice plant leaves, Yuan et al. \cite{b14} experimented with various color based features like normalized RGB, HSV, and Dark Green Color Index. Their objective was to establish a correlation between these features and the amount of chlorophyll present in the rice plant leaves with the help of a SPAD-502 meter. The SPAD-502 meter is a hand held device which can be used to get an accurate measure of leaf chlorophyll concentration. One of the setbacks of this method is that, the authors used only three positions to collect data for the SPAD-502 meter. If they had considered more number of positions then the measurement would have been more accurate. 

To assist human annotators quickly identify Nitrogen deficiencies in maize plants, Zermas et al. \cite{b10} proposed a solution based on Support Vector Machines(SVMs). Three different SVMs were used: the first one was used to separate green pixels from the rest (i.e.,  yellow and soil colored ones). The second one performed another classification to separate yellow pixels from the rest;  finally, the third SVM determined whether a particular pixel was yellow due to Nitrogen deficiency. The proposed approach was shown to be much faster than human annotators in creating image data sets for Nitrogen deficiency classification. Additionally, the authors presented an approach which utilized the dataset to build a model which determined whether there was Nitrogen deficiency in the images. 

Unfortunately, most of past published research work does not address the issue of early identification of abnormalities. Moreover, apart from a few examples, most of them focus on examining individual leaves or plants, instead of considering a larger area of plants. Data collection is also manual using hand-held instruments which limits generality and large scale deployment of the proposed approaches. Many previous methodologies are also abnormality-depended which limits their effectiveness to different types of abnormalities. Our goal is to develop a general-purpose abnormality detection system which leverages modern technologies for data collection and can detect different types of abnormalities as early as possible from a large area of plants of various growth stages. 

\section{Dataset}

We have experimented with a publicly available data set which contains images of maize plants of various growth stages exhibiting mostly Nitrogen deficiency but also abnormalities due to other types of nutrient deficiency and dryness \cite{b10}. The data set contains high resolution RGB images captured by a UAV at two different locations. For the purpose of this study, we used images from the V8 and V12 growth stages from the Becker field. Here, V stands for vegetative stage and the number associated with it represent how many leaves are present in the plant. It should be noted that although the labels of the data set indicate that the plants come from the V8 and V12 growth stages, there are several images in each category which contain maize plants from different growth stages. Moreover, although nutrient deficiencies typically start at an early growth stage, this is not always the case as we have confirmed from the images provided in the data set. The image resolution for V8 and V12 images are $4000\times6000\times3$ and $6000\times4000\times3 $ respectively. With the help of an annotation tool named Label Studio \cite{b15}, 44 images from the V8 stage and 46 images from the V12 stage were annotated using a bounding box to specify areas of abnormality (ground truth). Any yellow leaves on the ground were excluded from ground truth labeling.
\begin{figure}[t]
    \centering
    \subfigure[]{\includegraphics[width=0.49\textwidth]{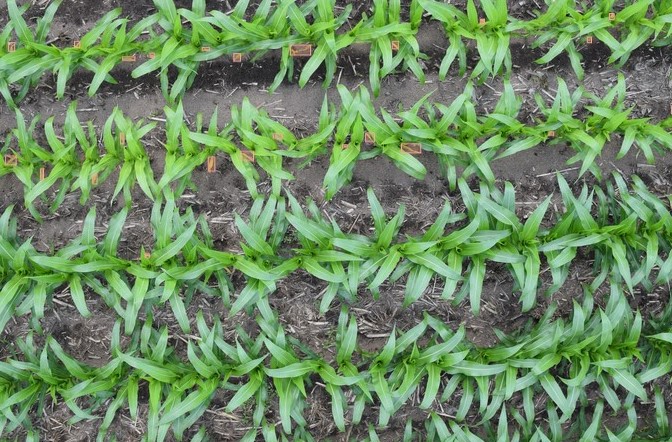}} 
    \subfigure[]{\includegraphics[width=0.49\textwidth]{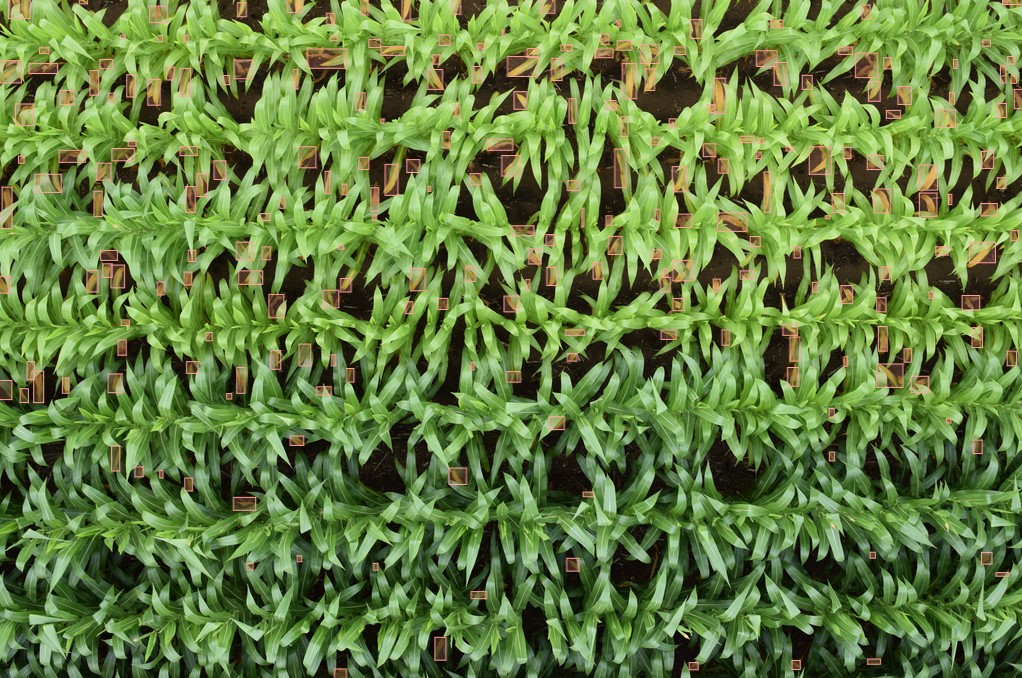}} 
    
    \caption{Illustration of abnormalities in large scale images for (a) V8 and (b) V12 growth stages.}
    
    \label{fig:large_scale_bounding}
\end{figure}

Figure \ref{fig:large_scale_bounding} shows two representative annotated images from the V8 and V12 growth stages. For training purposes, we selected 36 images from V8 and 36 images from V12 (we refer to this as set A); the rest, 8 images from V8 and 10 images from V12, were selected for testing purposes (we refer to this as set B). Based on the annotations, three separate data sets were created, one for training and two for testing. The training set was created by randomly cropping $250 \times 250 \times 3$ sub-images from set A as we wanted to build a general purpose abnormality detector which can perform well on any part of the field. To evaluate the performance of the abnormality detector (i.e., EfficientNet-B0), the first test data set was created in the same way (i.e., randomly cropped sub-images) using set B. To evaluate abnormality quantification, the second test data set was created from set B again but now using a non-overlapping sliding window approach to make sure that the whole image is covered.  

\begin{figure}[t]
    \centering
    \subfigure[]{\includegraphics[width=0.215\textwidth]{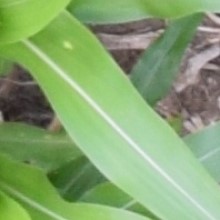}} 
    \subfigure[]{\includegraphics[width=0.22\textwidth]{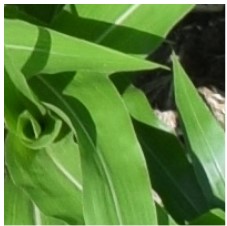}}
    \subfigure[]{\includegraphics[width=0.215\textwidth]{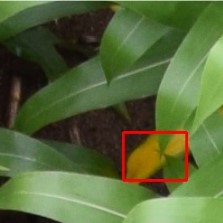}} 
    \subfigure[]{\includegraphics[width=0.216\textwidth]{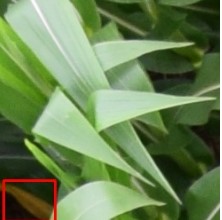}}
    
    \caption{Illustration of normal (a) \& (b)  and abnormal (c) \&(d) samples.}
    
    \label{fig:normal_abnormal}
\end{figure}

A particular sub-image was labelled as abnormal if it completely contained at least one bounding box. A total of 4966 randomly selected abnormal sub-images and 4966 randomly selected normal sub-images were extracted for training. Normal samples did not include any part of a bounding box. For validation purposes, 10\% of the training data was used. The first test set was built by selecting a total of 1211 abnormal and 1211 normal randomly cropped sub-images. Figure \ref{fig:normal_abnormal} shows some representative normal and abnormal sub-images. Since the UAV captures a big portion of the field which might contain 100-120 maize plants, it was determined that this area might be too big to be labeled as abnormal or normal. Therefore, each large scale test image was partitioned in 4 quarters, yielding 72 quarter scale images, each containing an average of 25-30 maize plants. The second test set was created from the quarter scale images using a sliding-window approach yielding a total of 6912 sub-images; each sub-image in this case was labelled as abnormal if it contained any part of a bounding box and normal otherwise.

\section{Methodology}
The proposed approach assumes that high resolution maize images have been collected by a UAV. The goal is to analyze each image collected to determine whether the plants exhibit low, medium, high or no abnormality. Our system does not currently perform any abnormality localization but we plan to include this feature in future versions of the system.  

Two approaches have been considered to quantify the level of abnormality in a test image. In the first approach, we consider all non-overlapping sliding windows in the image where each sliding window is classified as normal or abnormal using a deep learning classifier as discussed later. To quantify the level of abnormality in the whole image, we take the ratio of abnormal sliding windows over the total number of sliding windows; we refer to this ratio as the abnormal window probability and represents a coarse estimate of the amount of abnormality present in an image. We have empirically divided the window probability in several intervals as shown in Table \ref{tab:thresholds} in order to quantify the amount of abnormality as low, medium, high and no abnormality. In practice, these thresholds should be set up with the help of an expert.

\begin{figure}[t]
    \centering
    \subfigure[]{\includegraphics[width=0.25\textwidth]{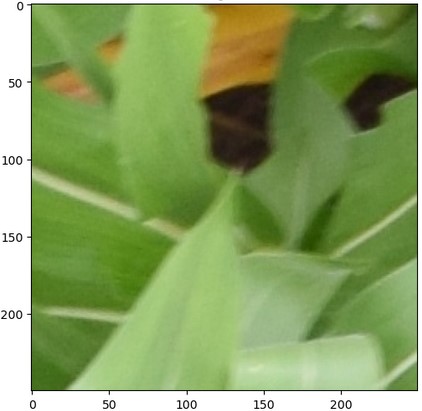}} 
    \subfigure[]{\includegraphics[width=0.25\textwidth]{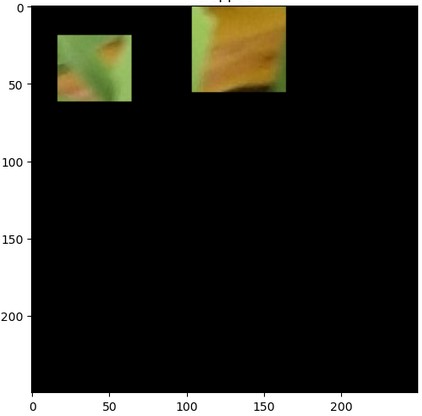}} 
    \subfigure[]{\includegraphics[width=0.25\textwidth]{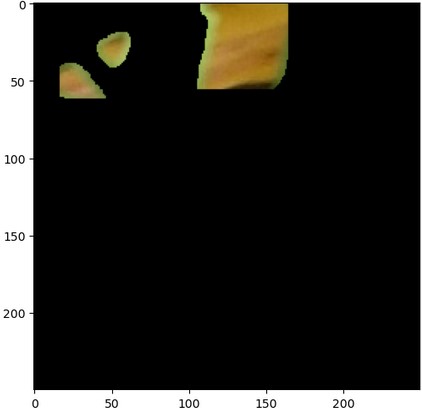}} 
    
    \caption{(a) A representative abnormal image (b) bounding boxes focusing on abnormality (c) abnormal pixels using color segmentation.}
    
    \label{fig:color_segmentation}
\end{figure}

In the second approach, we consider all non-overlapping sliding windows again in a test image and count all abnormal pixels in each sliding window based on the bounding boxes which are inside the window. We then add the abnormal pixels from all sliding windows and divide the sum by the total number of pixels in the image. We refer to this ratio as the abnormal pixel probability and represents a finer estimate of the amount of abnormality present in the image. Typically, computing the abnormal pixel probability requires segmenting the area of abnormality within the bounding boxes. Here, we have used a simple color segmentation scheme to roughly estimate the abnormal area within each bounding box. This was performed using simple thresholding in the HSV color space to extract yellow colored pixels inside any bounding boxes present in the image. Figure \ref{fig:color_segmentation} illustrates the color segmentation task for a representative abnormal image.

Using a Pearson correlation we verified a high positive correlation between the abnormal window and pixels probabilities. To quantify the amount of abnormality present in an image, we chose the thresholds for the abnormal pixel probability based on the thresholds of the abnormal window probability such that the number of test images in each category between the two approaches remains the same. However, there is no guarantee that the same exactly images belong to the same category for each method. This is because there are some borderline cases where the image is considered of low abnormality using the window probability but of medium abnormality using the pixel probability; this is also the case for the medium and high abnormality categories. Apart from this, there might be a bounding box in the computation of the window probability which falls into multiple neighboring sliding windows; as a result, the window probability can be overestimated. This is not the case when computing the abnormal pixel probability, however, the computation of the abnormal pixel probability suffers from possible segmentation errors due to using a rather simple color segmentation scheme. In the future, we plan to compute both probabilities more accurately. Table \ref{tab:thresholds} shows the corresponding thresholds for the abnormal pixel probability as well as the number of quarter scale test images in each category.
\begin{table}[t]
\centering
\caption{Quantification of different abnormalities (probability $x$ is scaled in the range [0-100] for both methods)}
\label{tab:thresholds}
\begin{tabular}{|c|c|c|c|}
\hline
 &
  \begin{tabular}[c]{@{}c@{}}Abnormal Window\\ Probability\end{tabular} &
  \begin{tabular}[c]{@{}c@{}}Abnormal Pixel\\ Probability\end{tabular} &
  \begin{tabular}[c]{@{}c@{}}No. of \\ Images\end{tabular} \\ \hline
None   & 0                           & 0                                 & 7  \\ \hline
Low    & 0\textless{}x\textless{}5    & 0\textless{}x\textless{}0.0415    & 9  \\ \hline
Medium & 5\textless{} x\textless{}20 & 0.0415\textless{} x\textless{}0.80 & 19 \\ \hline
High   & x\textgreater{} 20          & x\textgreater{} 0.80               & 37 \\ \hline
\end{tabular}
\end{table}
\subsection{Custom EfficientNet-B0 Classifier}

\begin{figure}[b]
    \centering
    \includegraphics[width=0.65\textwidth]{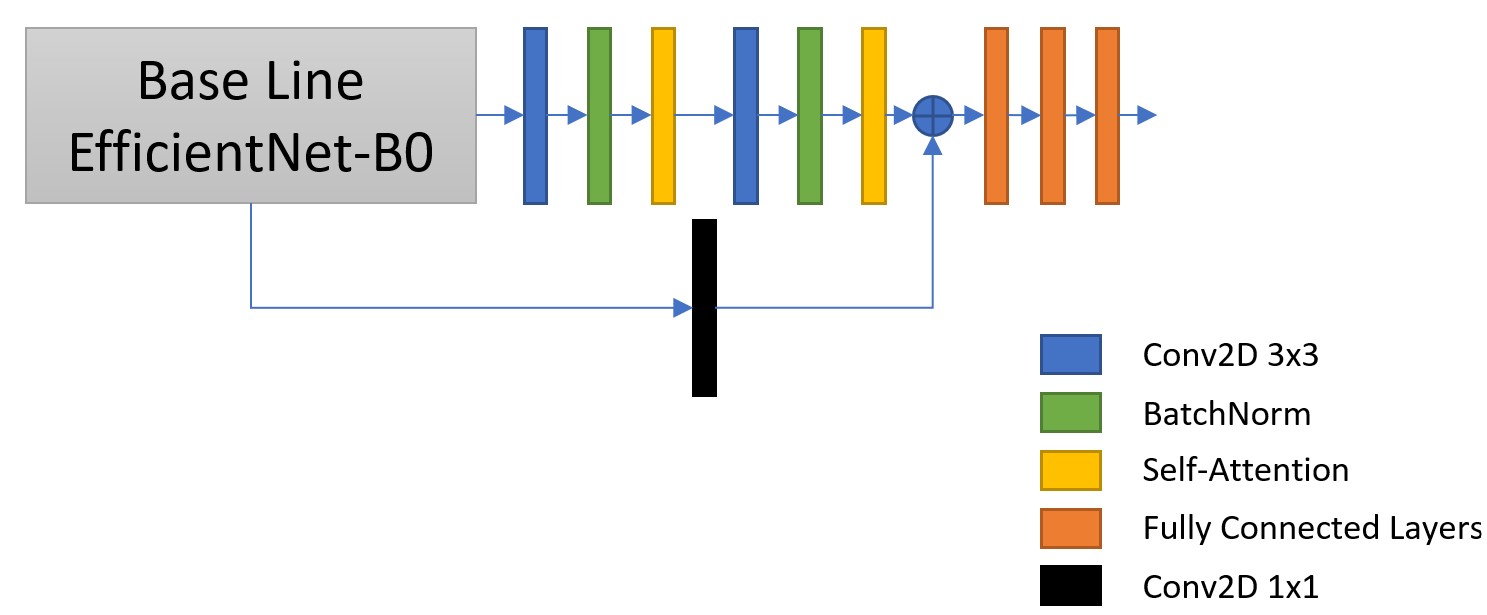} 
    \caption{Proposed customized EfficientNet-B0.}
    
    \label{fig:cus_ef}
\end{figure}

EfficientNet-B0 architecture is a light weight model which is capable of performing classification tasks very well which is the main reason for choosing this network in our study. Later versions of this network does perform slightly better but those require more time to train as they have significantly larger parameters. We have modified this model to better suit it in the context of our application. Generally, it is considered that any classification deep neural network has two parts. The first one is the feature extraction part which contains the convolutional, maxpooling, normalization layers etc. The second part is the classification layer which takes the features from the previous layers, performs average pooling, and feeds the results to a few fully-connected (FC) layers. Here, we have removed the classification part  and have inserted in its place several convolutional, batch normalization and self attention layers. The reason for including self-attention layers is that abnormalities occupy only a small area in the image; therefore, the model is expected to perform better if it focuses on a small region only. Additionally, a skip connection was introduced to retain both information and gradients that could potentially be lost during the training process. Following the inclusions of these layers, three FC layers we used for performing classification. Figure \ref{fig:cus_ef} provides a visual illustration of the customized model used in this study. 

\subsection{Abnormality Quantification Using Window Probability}
\begin{figure}[t]
    \centering
    \includegraphics[width=0.75\textwidth]{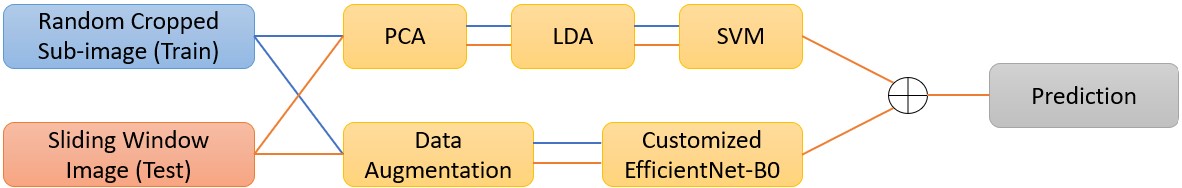}

    \caption{Training and testing phases for predicting the abnormal window probability.}
    
    \label{fig:methodology}
\end{figure}
To quantify the amount of abnormality in a test image using the window probability, each sliding window in the test image must be classified as normal or abnormal. We have experimented with two different classifiers and their fusion: the customized EfficientNet-B0 described earlier and a Support Vector Machine (SVM) classifier \cite{b16,b17}. The fusion model predicted an image as abnormal if either one of the above classifiers predicted it as abnormal. We decided to fuse a traditional machine learning model with a deep learning model since their solutions would be rather different which typically benefits fusion schemes most. Both the customized EfficientNet-B0 model and the SVM model were trained on the same training data and optimized using the validation data. During testing, each model was evaluated using the randomly cropped test set. As the training and test images were $250\times250\times3$, Principal Component Analysis(PCA) followed by Linear Discriminant Analysis (LDA) were performed in the case of SVM to help it find more powerful features. For the deep learning model, various data augmentation approaches (e.g., random vertical and horizontal flips) were performed. After the classification of each sliding window as normal or abnormal, the window probability was computed and the test image was assigned to an abnormality category or to the normal category using the thresholds shown in Table \ref{tab:thresholds}. Finally, the accuracy was calculated based on the original and predicted labels. An illustration of the training and testing phases is shown in Figure \ref{fig:methodology}.
\subsection{Abnormal Quantification Using Pixel Probability}
In order to predict the abnormal pixel probability of a test image, a regression model, namely Histogram-based Gradient Boosting Regression Tree, was applied to each sliding window in the test image \cite{b18}. This is a faster version of the Gradient Boosting Regression Tree \cite{b19}. The regression model was trained on the randomly cropped abnormal samples only (i.e., it wasn't trained on the whole data set because of the zero inflation problem\cite{b22}). Since the dimension of each sliding window is $250\times250\times3$, which is rather big for the regression model, PCA was used to reduce the dimension by preserving 99\% of the variance in the data, leading to 3394 features. The abnormal pixel probability for each small scale test image was computed by summing up the predicted pixel probabilities for all sliding windows. Each test image was then assigned to the appropriate abnormal category using the thresholds shown in Table \ref{tab:thresholds}. Finally, the accuracy was calculated based on the original and predicted labels.

\section{Results and Discussion}

In this section a short description about the parameters of the models and results obtained from the experimentations are discussed. 

\subsection{Experimental Setup}

The randomly cropped data sets described in Section 3 were used to train and test the classification and regression models. EfficientNet-B0 was trained for 200 epochs using SGD optimizer with a learning rate of 0.0001. Additionally, one-cyclic learning rate was used for updating the learning rate in each epoch \cite{b20}. In order to train the SVM model, we used an RBF kernel; the value of C was set to 1. The regression model was trained for 750 iterations using the squared loss, a max depth value of 7, and the L2 regularization value set to 3. Other parameter values were experimented as well but for these values provided the best results. 

\subsection{Abnormal vs Normal Classification}
\begin{table}[t]
\centering
\caption{Performance comparison of different models}
\label{tab:model_comparison}
\begin{tabular}{|c|c|c|c|l}
\cline{1-4}
 &
  \begin{tabular}[c]{@{}c@{}}Training\\ Accuracy(\%)\end{tabular} &
  \begin{tabular}[c]{@{}c@{}}Validation\\ Accuracy(\%)\end{tabular} &
  \begin{tabular}[c]{@{}c@{}}Test\\ Accuracy(\%)\end{tabular} &
   \\ \cline{1-4}
ResNet152                                                            & 93.75 & 94.06 & 92.36 &  \\ \cline{1-4}
DenseNet201                                                          & 95.00 & 93.06 & 92.90 &  \\ \cline{1-4}
EfficientNet-B0                                                      & 94.69 & 93.96 & 92.69 &  \\ \cline{1-4}
\begin{tabular}[c]{@{}c@{}}Customized EfficientNet-B0\end{tabular} & 99.97 & 93.26 & \textbf{93.15} &  \\ \cline{1-4}
\end{tabular}
\end{table}

Table \ref{tab:model_comparison} compares the performance of the customized EfficientNet-B0 with the original EfficientNet-B0 as well as with DenseNet-201 and ResNet-152 using the randomly cropped data set. As it can be observed from the results, the customized EfficientNet-B0 model outperformed all other models, including the original EfficientNet-B0 model. 

\subsection{Results Based on Abnormal Window Probability}

Table \ref{tab:ab_sl_w_results_w_prob} shows the results obtained on the small scale test images. The rows of the table show the performance achieved for the normal and abnormal categories while the columns correspond to the SVM, Customized EfficientNet-B0 mdoels and their fusion. As shown in the table, the model based on fusion out-performs the other two models.  
\begin{table}[t]
\centering
\caption{Performance on sliding window test set for abnormal window probability}
\label{tab:ab_sl_w_results_w_prob}
\begin{tabular}{|c|c|c|c|}
\hline
 & SVM(\%) & \begin{tabular}[c]{@{}c@{}}Customized\\ EfficientNet-B0(\%)\end{tabular} & \begin{tabular}[c]{@{}c@{}}Fusion \\ Model(\%)\end{tabular} \\ \hline
None   & 100   & 100   & 100   \\ \hline
Low    & 55.56 & 77.78 & 88.89 \\ \hline
Medium & 63.16 & 68.42 & 84.21 \\ \hline
High   & 51.35 & 94.59 & 100   \\ \hline
\end{tabular}
\end{table}

\begin{figure}[t]
    \centering
    \subfigure[]{\includegraphics[width=0.40\textwidth]{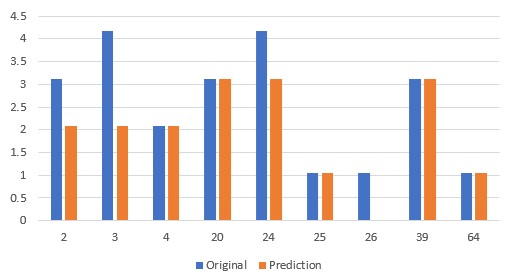}} 
    \subfigure[]{\includegraphics[width=0.50\textwidth]{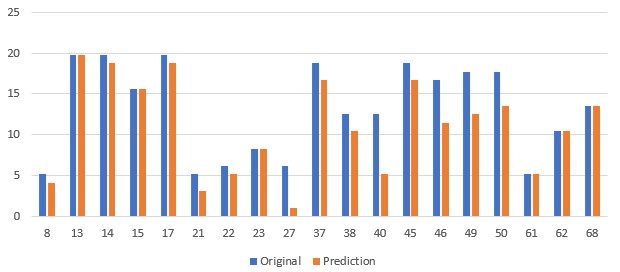}} 
    \subfigure[]{\includegraphics[width=0.90\textwidth]{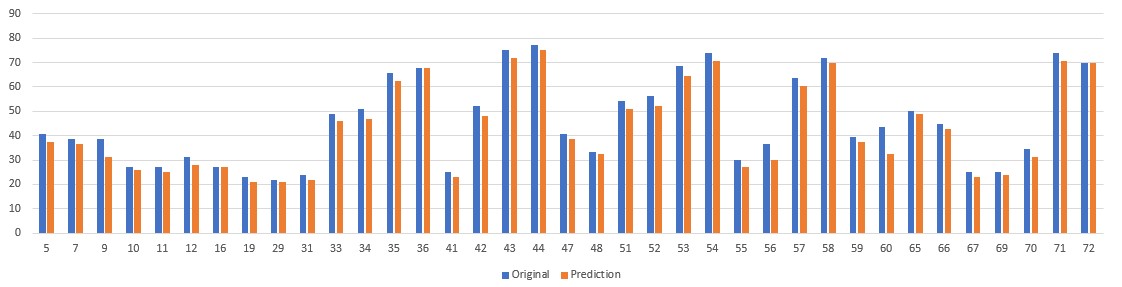}}
    
    \caption{Performance comparison between the original and predicted window probabilities for (a) low (b) medium and (c) high abnormal categories.}
    
    \label{fig:ab_win_category}
\end{figure}
\begin{figure}[t]
    \centering
    \subfigure[]{\includegraphics[width=0.23\textwidth]{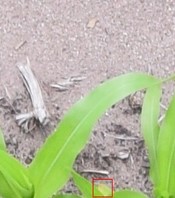}} 
    \subfigure[]{\includegraphics[width=0.249\textwidth]{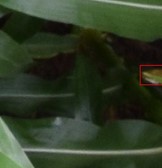}} 
    \subfigure[]{\includegraphics[width=0.30\textwidth]{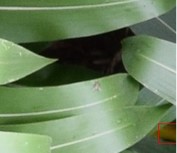}}
    
    \caption{Sample images that are classified wrongly by the classifier.}
    
    \label{fig:misclf_images}
\end{figure}

To better understand the fusion model (best model), a bar plot was created for each abnormal category (see Figure \ref{fig:ab_win_category}). The bar plot depicts a side by side comparison between the original and predicted window probability values for each image in each category. In the plots, the x-axis represents the ID numbers of the individual test images and the y-axis represents the window probability (scaled in the interval [1-100]).  The model miss-classified 1 out of 9 test images and 3 out of 19 images for low and medium abnormality category. For the low abnormality category, the model was not able to correctly predict any abnormal window for a particular test image. A detailed analysis revealed that the particular test image (i.e., \#26) had only one abnormal sliding window out of 96 sliding windows. That particular window had a very small amount of abnormal pixels. It appears that this window was too difficult for the model to be classified correctly. In case of the medium abnormality category, it can be observed from Figure \ref{fig:ab_win_category}(b) that the test images miss-classified (i.e., images \#8, \#21 and \#61) had abnormal window probability values slightly over 5\% which was the threshold between the low and medium abnormality categories. For hard thresholds, like in this study, the number of similar errors would increase with more categories. Figure \ref{fig:misclf_images} shows several test images which have been labelled as normal by the classifier but are in fact abnormal images as shown by the red bounding box. As it can be observed, occlusion, lighting, small region of interest are some of the factors that hamper the performance of the model. 

\subsection{Results Based on Abnormal Pixel Probability}
With the help of the regression model 100\%, 66.67\%, 100\% and 89.19\% accuracy for the abnormal pixel probability was obtained for None, Low, Medium and High category through the experimentations. As it can be observed, the regression model performs very well for normal and medium abnormality category; however, its performance is lower on the low and high abnormality categories.
\begin{figure}[!h]
    \centering
    \subfigure[]{\includegraphics[width=0.38\textwidth]{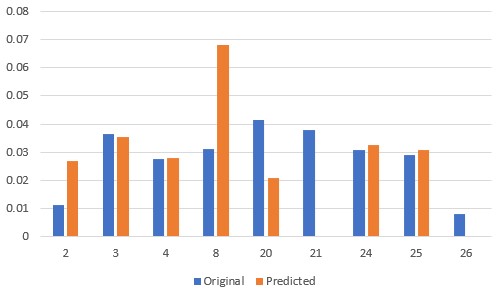}} 
    \subfigure[]{\includegraphics[width=0.48\textwidth]{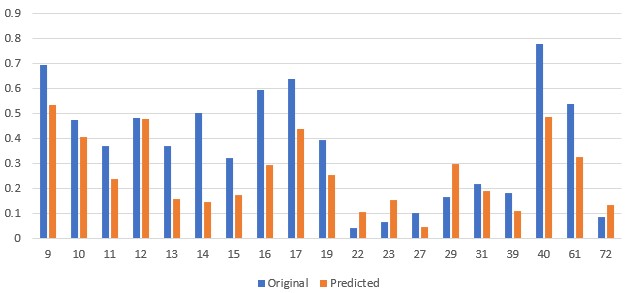}} 
    \subfigure[]{\includegraphics[width=0.88\textwidth]{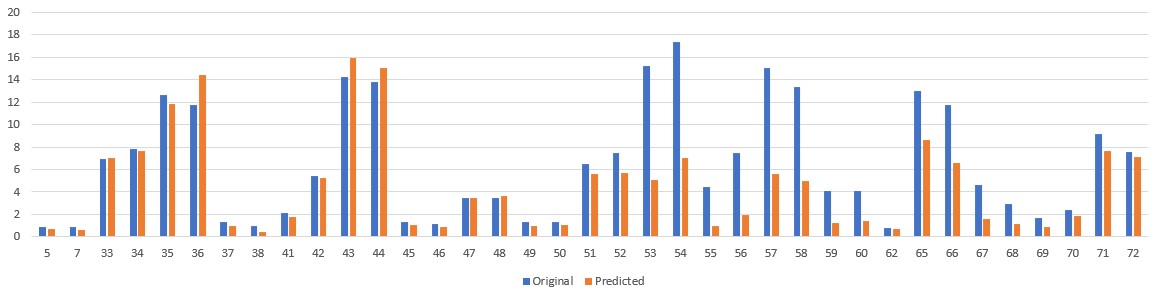}}
    
    \caption{Performance comparison between the original value and predicted value for (a) low (b) medium and (c) high abnormal pixel probability.}
    
    \label{fig:ab_pixel_category}
\end{figure}
A bar plot was created again to better understand the results obtained in this case (see Figure \ref{fig:ab_pixel_category}). As it can be observed, the regression model predicted a zero abnormal pixel probability for two images in this category (i.e., images \#21 and \#26). Image \#26 is the same image mistaken by the window probability approach. A detailed analysis on image \#21 revealed that there were only two windows which contained abnormalities and due to illumination and occlusion issues the model was not able to provide any regression values. There were also four test images which were not correctly categorized in the high abnormality category(i.e., images \#5,\#7\#38 and \#62). These had predicted values which were very close to the threshold while the original values were just above the threshold for the high abnormality category. There are more factor that have contributed to errors using abnormal pixel probability such as inaccurate estimation of these probabilities in the training set and dried yellowish leaves on the ground.

\section{Conclusions}
Detection of abnormalities in maize plants in early stages is extremely crucial. However, even in cases when the abnormality has progressed to some extend, certain actions can still be taken in order to ensure proper growth and yields. This research work focused on identifying abnormalities using UAV images to detect whether a particular area contains low, medium, high or no abnormality which mitigates the aforementioned problem. Abnormal window probability and abnormal pixel probability approaches were considered to quantify potential abnormalities in the field. We have reported promising preliminary results using a publicly available data set. We plan to estimate the pixel probability more accurately as well as fuse the window and pixel probabilities to improve abnormality quantification. Since the original data was not collected by us we could not recommend ideal UAV heights, camera view, optimal illumination conditions etc. In the future, we plan to experiment with more data including data from different growth stages, different locations, and exhibiting different types of abnormalities. Finally, we plan to incorporate abnormality localization capabilities to better assist human annotators to create larger data sets for training deep learning models.   

\textbf{Acknowledgement}: This work was supported by the National Institute of Food and Agriculture/USDA, Award No. 2020-67021-30754.


\begin{thebibliography}{8}

\bibitem{b1}

Du, H. et al. ``The elements of end-to-end deep face recognition: A survey of recent advances". ACM Computing Surveys (CSUR), 54(10s), 1-42, (2022).

\bibitem{b2}

Hao, T. et al. ``Recent progress in leveraging deep learning methods for question answering". Neural Computing and Applications, 1-19, (2022). 

\bibitem{b3}

He, K. et al. ``Transformers in medical image analysis: A review". Intelligent Medicine. (2022).

\bibitem{b4}
Fendji, J. et al. ``Automatic speech recognition using limited vocabulary: A survey". Applied Artificial Intelligence, 36(1), 2095039, (2022).

\bibitem{b5}
Shaikh, T. A., Rasool, T., \& Lone, F. R.  ``Towards leveraging the role of machine learning and artificial intelligence in precision agriculture and smart farming". Computers and Electronics in Agriculture, 198, 107119.(2022).

\bibitem{b6}
Coulibaly, S. et al. ``Deep learning for precision agriculture: A bibliometric analysis". Intelligent Systems with Applications, 16, 200102.
(2022).

\bibitem{b7}
Nseir, A. and  Honig, L.. ``Corn and Soybean Production down in 2022, USDA Reports Corn Stocks down, Soybean Stocks down from Year Earlier Winter Wheat Seedings up for 2023",United States Department of Agriculture. 12 Jan. 2023, www.nass.usda.gov/Newsroom/2023/01-12-2023.php. 

\bibitem{b8}
Barbedo A., Garcia J. ``Digital image processing techniques for detecting, quantifying and classifying plant diseases". SpringerPlus, 2(1), 1-12. (2013).

\bibitem{b9}
Romualdo, L. M. et al. ``Use of artificial vision techniques for diagnostic of nitrogen nutritional status in maize plants". Computers and electronics in agriculture, 104, 63-70.(2014).

\bibitem{b21}
Sethy, P. K. et al. ``Nitrogen deficiency prediction of rice crop based on convolutional neural network". Journal of Ambient Intelligence and Humanized Computing, 11, 5703-5711.(2020). 

\bibitem{b10}
Zermas D. et al. ``A Methodology for the Detection of Nitrogen Deficiency in Corn Fields Using High-Resolution RGB Imagery,"  IEEE Transactions on Automation Science and Engineering, vol. 18, no. 4, pp. 1879-1891, Oct. (2021).

\bibitem{b11}
Chore, A., \& Thankachan, D.  ``Nutrient Defect Detection In Plant Leaf Imaging Analysis Using Incremental Learning Approach With Multifrequency Visible Light Approach". Journal of Electrical Engineering \& Technology, 18(2), 1369-1387. (2023).

\bibitem{b12}
 Rahadiyan, D. et al., ``Classification of Chili Plant Condition based on Color and Texture Features"  Seventh International Conference on Informatics and Computing (ICIC), Denpasar, Bali, Indonesia, pp. 01-07. (2022).

\bibitem{b13}
 N, Tejasri. et al., ``Drought Stress Segmentation on Drone captured Maize using Ensemble U-Net framework," 2022 IEEE 5th International Conference on Image Processing Applications and Systems (IPAS), Genova, Italy,  pp. 1-6, (2022).
 
\bibitem{b14}
Yuan, Y., et al. ``Diagnosis of nitrogen nutrition of rice based on image processing of visible light" 2016 IEEE International conference on functional-structural plant growth modeling, simulation, visualization and applications (FSPMA) (pp. 228-232). (2016).


\bibitem{b15}
Tkachenko, M., et al. ``Label Studio: Data labeling software" url: https://github.com/heartexlabs/label-studio. (2020-22).

\bibitem{b16}
Cortes, C., \& Vapnik, V. ``Support-vector networks". Machine learning, Springer 20, 273-297. (1995).

\bibitem{b17}
Tan, M., \& Le, Q. ``Efficientnet: Rethinking model scaling for convolutional neural networks". In International conference on machine learning. PMLR. (pp. 6105-6114). (2019).


\bibitem{b18}
Ke, G. et al. ``Lightgbm: A highly efficient gradient boosting decision tree". Advances in neural information processing systems, vol 30. (2017).

\bibitem{b19}
Friedman, J. H. ``Stochastic gradient boosting". Computational statistics \& data analysis, 38(4), 367-378. (2002).

\bibitem{b22}
Heilbron, D. C. ``Zero‐altered and other regression models for count data with added zeros". Biometrical Journal, 36(5), 531-547.(1994).

\bibitem{b20}
Smith, L. N., \& Topin, N. ``Super-convergence: Very fast training of neural networks using large learning rates". Artificial intelligence and machine learning for multi-domain operations applications. Vol. 11006, pp. (369-386). (2019).



\end{thebibliography}
\end{document}